\documentclass[sigconf,natbib=true,anonymous=false]{acmart}

\AtBeginDocument{%
}

\copyrightyear{2026}
\acmYear{2026}
\setcopyright{cc}
\setcctype{by}
\acmConference[SIGIR '26]{Proceedings of the 49th International ACM SIGIR Conference on Research and Development in Information Retrieval}{July 20--24, 2026}{Melbourne, VIC, Australia}
\acmBooktitle{Proceedings of the 49th International ACM SIGIR Conference on Research and Development in Information Retrieval (SIGIR '26), July 20--24, 2026, Melbourne, VIC, Australia}
\acmDOI{10.1145/3805712.3809863}
\acmISBN{979-8-4007-2599-9/2026/07}

\makeatletter
\gdef\@copyrightpermission{
\begin{minipage}{0.3\columnwidth}
\href{https://creativecommons.org/licenses/by/4.0/}{\includegraphics[width=0.90\textwidth]{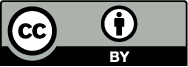}}
\end{minipage}\hfill
\begin{minipage}{0.7\columnwidth}
\href{https://creativecommons.org/licenses/by/4.0/}{This work is licensed under a Creative Commons Attribution International 4.0 License.}
\end{minipage}
\vspace{5pt}
}
\makeatother

\usepackage{enumitem}
\usepackage{xspace}
\usepackage{subcaption}
\usepackage{multirow}
\usepackage{colortbl}
\usepackage{array}
\usepackage{booktabs}
\newcolumntype{C}[1]{>{\centering\arraybackslash}p{#1}}
\usepackage{xcolor}
\usepackage{soul}
\usepackage[normalem]{ulem}

\usepackage{xurl} 

\usepackage{microtype}


\begin{document}



\title[From Raw Features to Effective Embeddings: A Three-Stage Approach for Multimodal Recipe Recommendation]{From Raw Features to Effective Embeddings: A Three-Stage Approach for Multimodal Recipe Recommendation}


\settopmatter{authorsperrow=3}
        \author{Jeeho Shin}
        \orcid{0009-0006-9202-916X}
	\affiliation{%
		\institution{KAIST}
            \city{Daejeon}
            \country{Republic of Korea}
	}
	\email{imhappy68@kaist.ac.kr}

        \author{Kyungho Kim} 
        \orcid{0009-0008-8304-9585}
	\affiliation{%
    	\institution{KAIST}
            \city{Seoul}
            \country{Republic of Korea}
	}
	\email{kkyungho@kaist.ac.kr}
	
	\author{Kijung Shin}
        \orcid{0000-0002-2872-1526}
	\affiliation{%
		\institution{KAIST}
            \city{Seoul}
            \country{Republic of Korea}
	}
	\email{kijungs@kaist.ac.kr}
    
\begin{abstract}

Recipe recommendation has become an essential task in web-based food platforms. 
A central challenge is effectively leveraging rich multimodal features beyond user–recipe interactions. Our analysis shows that even simple uses of multimodal signals yield competitive performance, suggesting that systematic enhancement of these signals is highly promising.
We propose \method, a 3-stage framework for recipe recommendation that progressively refines raw multimodal features into effective embeddings through:
(1) \textbf{content-based enhancement} using foundation models with multimodal comprehension,
(2) \textbf{relation-based enhancement} via message propagation over user–recipe interactions, and
(3) \textbf{learning-based enhancement} through contrastive learning with learnable embeddings.
Experiments on two real-world datasets show that \method outperforms existing methods, achieving 7–15\% higher Recall@10. 





\end{abstract}

\newcommand{\smallsection}[1]{\vspace{0.2pt}{\noindent {\bf \uline{\smash{#1}}}}}

\newtheorem{obs}{\textbf{Observation}}
\newtheorem{prp}{\textbf{Property}}
\newtheorem{dfn}{\textbf{Definition}}
\newtheorem{trm}{\textbf{Theorem}}

\newcommand\red[1]{\textcolor{red}{#1}}
\newcommand\blue[1]{\textcolor{blue}{#1}}
\newcommand\orange[1]{\textcolor{orange}{#1}}
\newcommand\brown[1]{\textcolor{brown}{#1}}
\newcommand\olive[1]{\textcolor{olive}{#1}}
\newcommand\sunwoo[1]{\textcolor{sunwooblue}{#1}}

\definecolor{mygreen}{rgb}{0,0.7,0}
\definecolor{sunwooblue}{RGB}{66, 133, 244}
\newcommand\green[1]{\textcolor{mygreen}{#1}}

\newcommand{\method}{\textsc{TESMR}\xspace}

\definecolor{verylightgray}{gray}{0.9}

\newcommand{\appropto}{\mathrel{\vcenter{
  \offinterlineskip\halign{\hfil$##$\cr
    \propto\cr\noalign{\kern1pt}\sim\cr\noalign{\kern-1pt}}}}}

\setlength{\textfloatsep}{0.10cm}
\setlength{\dbltextfloatsep}{0.10cm}
\setlength{\abovecaptionskip}{0.10cm}
\setlength{\skip\footins}{0.10cm}

\maketitle

\section{Introduction}
\label{sec:intro}
Recipe recommendation~\cite{zhang2024healthrec, zhang2024clussl, gao2022fgcn, song2023SCHGN, fu2025ChefMind} aims to suggest recipes that a user is likely to prefer, and is essential in web-based food platforms, including cooking websites and meal-planning services.

A key to effective recipe recommendation is leveraging rich multimodal features beyond user–recipe interactions. Recipes contain textual information such as titles and ingredient lists, as well as visual cues from food images. Users also provide textual reviews that directly reflect their preferences.

The importance of multimodal features is supported by our empirical analysis. We find that, even without training, a simple use of these signals already yields competitive performance (see Section~\ref{sec:analysis}).
This suggests that systematically enhancing multimodal features has the potential to substantially outperform existing methods.

Motivated by this finding, we propose \underline{t}hree \underline{e}nhancement \underline{s}teps for \underline{m}ultimodal \underline{r}ecipe recommendation (\method).
Our framework begins with content-based enhancement, where foundation models leverage their contextual understanding to distill raw data into concise yet informative summaries of users and recipes.
It then performs relation-based enhancement by encoding these summaries into initial embeddings and propagating them over the user–recipe interaction graph to incorporate collaborative signals. Finally, learning-based enhancement fuses multimodal and learnable embeddings through cross-view contrastive learning.

Our contributions are summarized as follows:
\begin{itemize}[leftmargin=*]
    \item \textbf{Observation.} 
    We show the importance and substantial potential of leveraging multimodal features in recipe recommendation.
    \item \textbf{Method.} We propose \textbf{\method}, a novel recipe recommendation framework with 3-stage enhancement of multimodal features.
    \item \textbf{Experiments.} We show experimentally that \textbf{\method} achieves 7–15\% higher Recall@10 compared to state-of-the-art baselines.
\end{itemize}
For \textbf{reproducibility}, our code and datasets are available at \url{https://github.com/JHshin6688/TESMR}.

\section{Related Work \& Problem Formulation}
\label{sec:prelim}

\smallsection{Related work on multimodal recommendation.}
Existing multimodal recommendation models~\cite{zhou2023FREEDOM, zhou2023BM3, zhou2023DRAGON, su2024SOIL}
have primarily focused on leveraging item-side multimodal information, such as titles, descriptions, and visual images. For example, FREEDOM~\cite{zhou2023FREEDOM} and DRAGON~\cite{zhou2023DRAGON} employ graph neural networks (GNNs) on user-item and item-item graphs to better model item relationships based on multimodal representations. Recently, foundation models have been widely applied to multimodal recommendation; see a survey~\cite{lopez2025survey}. Many of these works use foundation models as scorers or rankers, while we use them to enhance user and item features for GNNs, combining the strengths of both approaches.
Closely related to our work, 
Dang et al.~\cite{dang2025MLLMRec} and Meng et al.~\cite{meng2025doge} leverage foundation models to enrich item descriptions. With a focus on recipe recommendation, our approach applies foundation models to user reviews and their associated recipe summaries, constructing enriched user summaries in addition to enhanced recipe descriptions.

\smallsection{Related work on recipe recommendation.}
In recipe recommendation, we observe similar trends. Existing methods~\cite{gao2022fgcn, zhang2024clussl, song2023SCHGN, zhang2024healthrec} primarily focus on leveraging item-side (i.e., recipe) multimodal features, such as titles or recipe descriptions, often through propagation over recipe-ingredient or recipe-recipe graphs. 
Recent works~\cite{fu2025ChefMind, zhang2025MOPI-HFRS} enhance the interpretability of recommendations by using the reasoning capabilities of foundation models.
However, the use of foundation models for multimodal feature enhancement remains underexplored in the recipe recommendation literature.

\smallsection{Notations and Problem Formulation.}
Let $\mathcal{U}=\{u_{1},\dots,u_{\vert \mathcal{U} \vert}\}$, $\mathcal{R}=\{r_{1},\dots,r_{\vert \mathcal{R} \vert}\}$ denote the sets of users and recipes, respectively. We define the user-recipe interaction matrix $\mathbf{A} \in \{0,1\}^{\vert \mathcal{U} \vert \times \vert \mathcal{R} \vert}$, where $\mathbf{A}_{ur} = 1$ if user $u$ has interacted with recipe $r$ (e.g., by reviewing it), and $0$ otherwise.
Each recipe $r$ has a raw textual description $\mathbf{T}_{r}$ and an image $\mathbf{V}_{r}$. For each interaction, user $u$ provides a review $\mathbf{T}_{u}^{r}$ for recipe $r$. The goal of recipe recommendation is to predict future user-recipe interactions by leveraging both the interaction matrix ($\mathbf{A}$) and the multimodal information ($\mathbf{T}_{r}$, $\mathbf{V}_{r}$, and $\mathbf{T}_{u}^{r}$).

\section{Preliminary Data Analysis}
\label{sec:analysis}

In this section, we present our preliminary analysis using two real-world recipe recommendation datasets (Allrecipes\footnote{\url{https://www.kaggle.com/datasets/elisaxxygao/foodrecsysv1}} and Food.com\footnote{\url{https://www.kaggle.com/datasets/shuyangli94/food-com-recipes-and-user-interactions}}). Their statistics are provided in Table~\ref{tab:dataset}.

\begin{figure}[t]
    \centering
    \includegraphics[width=0.99\linewidth]
        {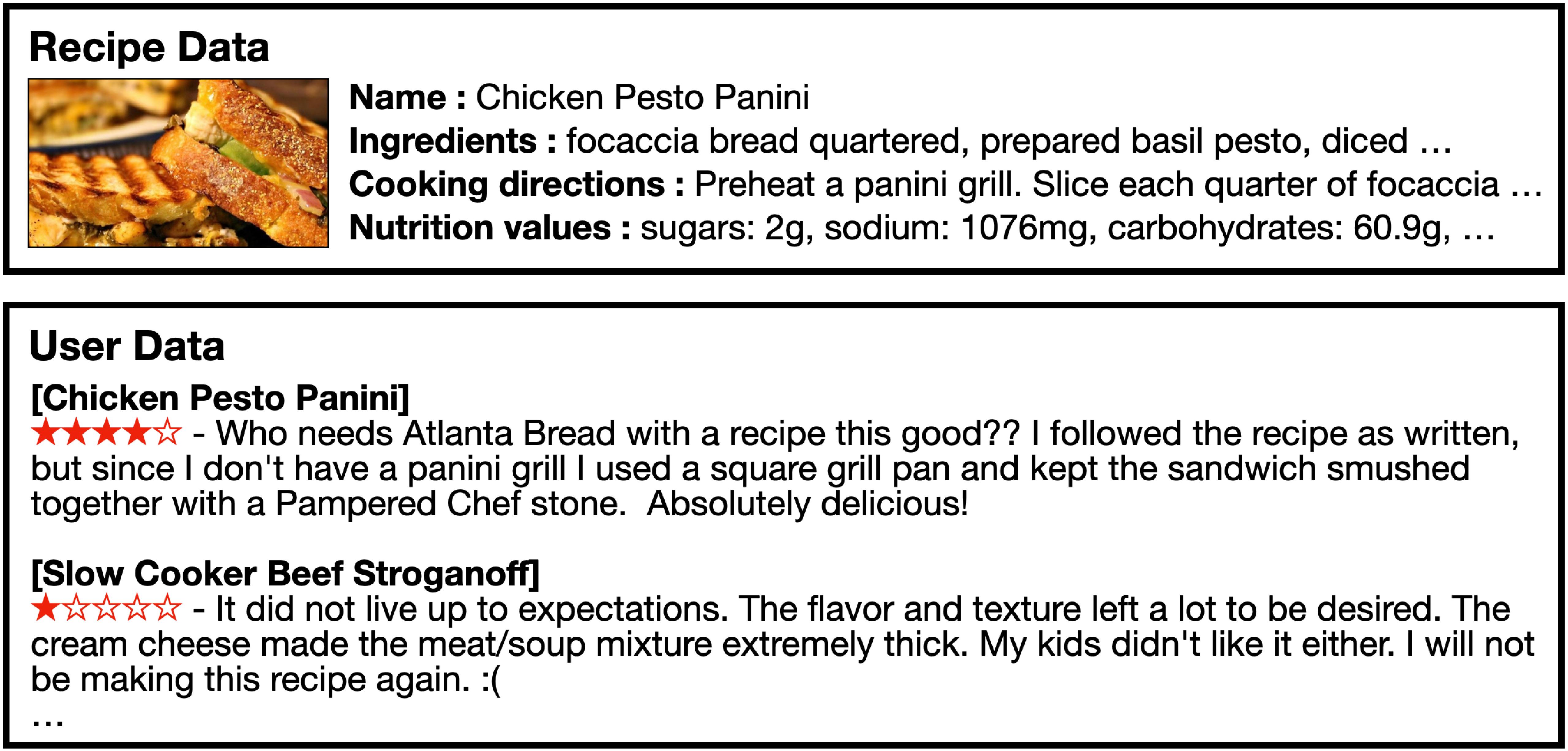}%
        \caption{Example multimodal data for recipes and users.\label{fig:data_example}}
\end{figure}

\begin{figure}[t]
    \vspace{-0.5mm}
    \centering        \includegraphics[width=0.99\linewidth]{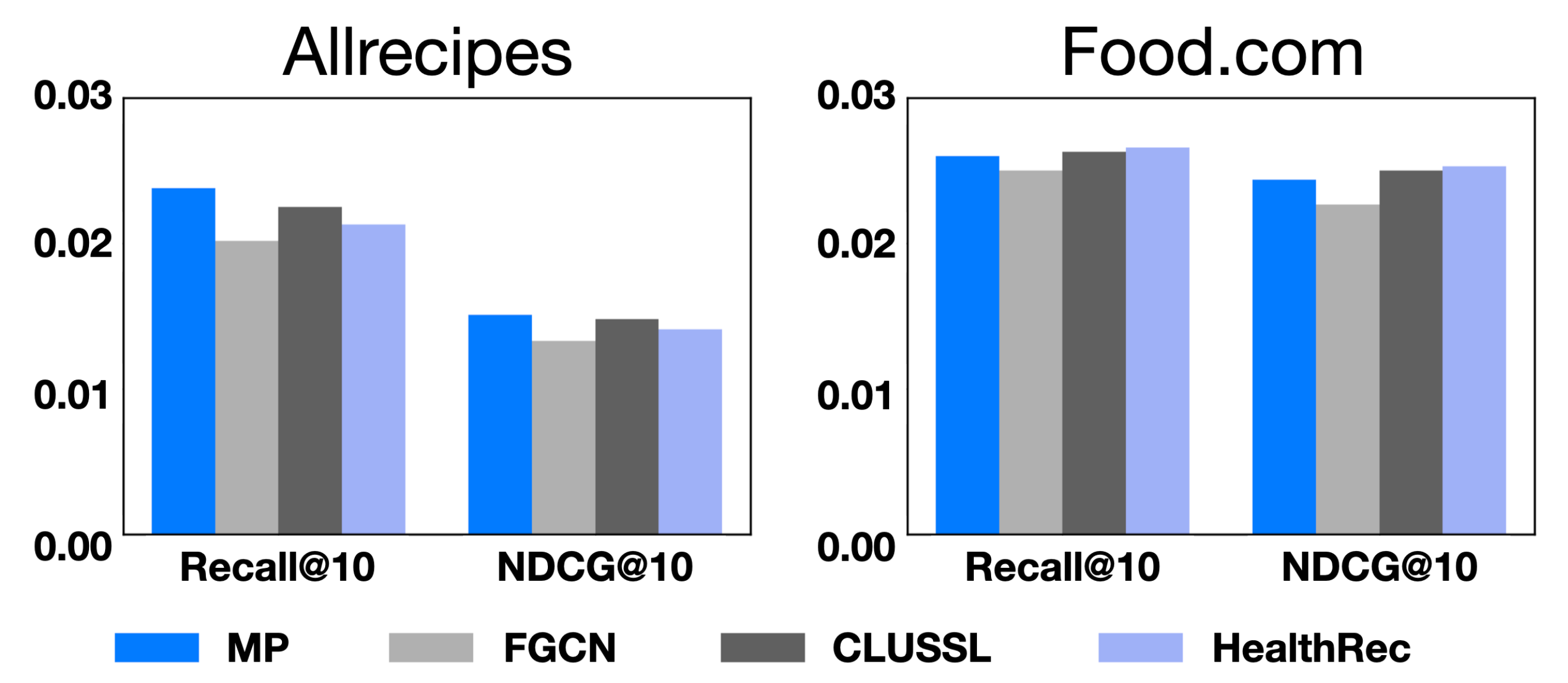}
        \caption{Observation: A simple use of multimodal features without training (MP) matches state-of-the-art performance.}
        \label{fig:data_obs}
\end{figure}

\smallsection{Setups.}
We apply a pretrained encoder\footnote{\url{https://huggingface.co/sentence-transformers/all-MiniLM-L6-v2}} to raw
features, illustrated in Figure~\ref{fig:data_example}. Specifically, for each recipe, we encode its textual content (names, ingredients, directions, and nutrition), and for each user, we encode their reviews with the corresponding recipe descriptions. We then perform message propagation on user-recipe interactions using the encoded representations, without any parameter updates, as in \cite{he2020lightgcn}.
Lastly, we compute the final scores for user–recipe pairs using the inner product of their representations.
\smallsection{Results.}
As shown in Figure~\ref{fig:data_obs}, even without any additional training or learnable parameters, this simple approach, denoted by MP, already performs comparably to or even outperforms several state-of-the-art models (FGCN~\cite{gao2022fgcn}, CLUSSL~\cite{zhang2024clussl}, and HealthRec~\cite{zhang2024healthrec}).
This highlights the importance of systematically leveraging multimodal features and suggests its potential for achieving substantial gains over state-of-the-art methods.

{\renewcommand{\arraystretch}{0.9}
\begin{table}[t]
\vspace{-4mm}
\begin{center}
\setlength\tabcolsep{3pt} 
\caption{\label{tab:dataset} Data statistics of real-world datasets.}
\scalebox{0.88}{
\begin{tabular}{l|cccccc}
    \toprule
    \textbf{Dataset} & \textbf{\#User} & \textbf{\#Recipe} & \textbf{\#Interaction} & \textbf{\#Ingredient} & \textbf{\#Sparsity (\%)}\\
    \midrule
    Allrecipes & 68,768 & 45,630 & 1,093,845 & 19,987 & 99.96\\
    Food.com & 7,585 & 29,905 & 322,546 & 4,963 & 99.86 \\
    \bottomrule
\end{tabular}}
\end{center}
\end{table}}

\section{Proposed Method: \method}
\label{sec:method}

\begin{figure*}[t]
    \centering
    \includegraphics[width=0.99\linewidth]{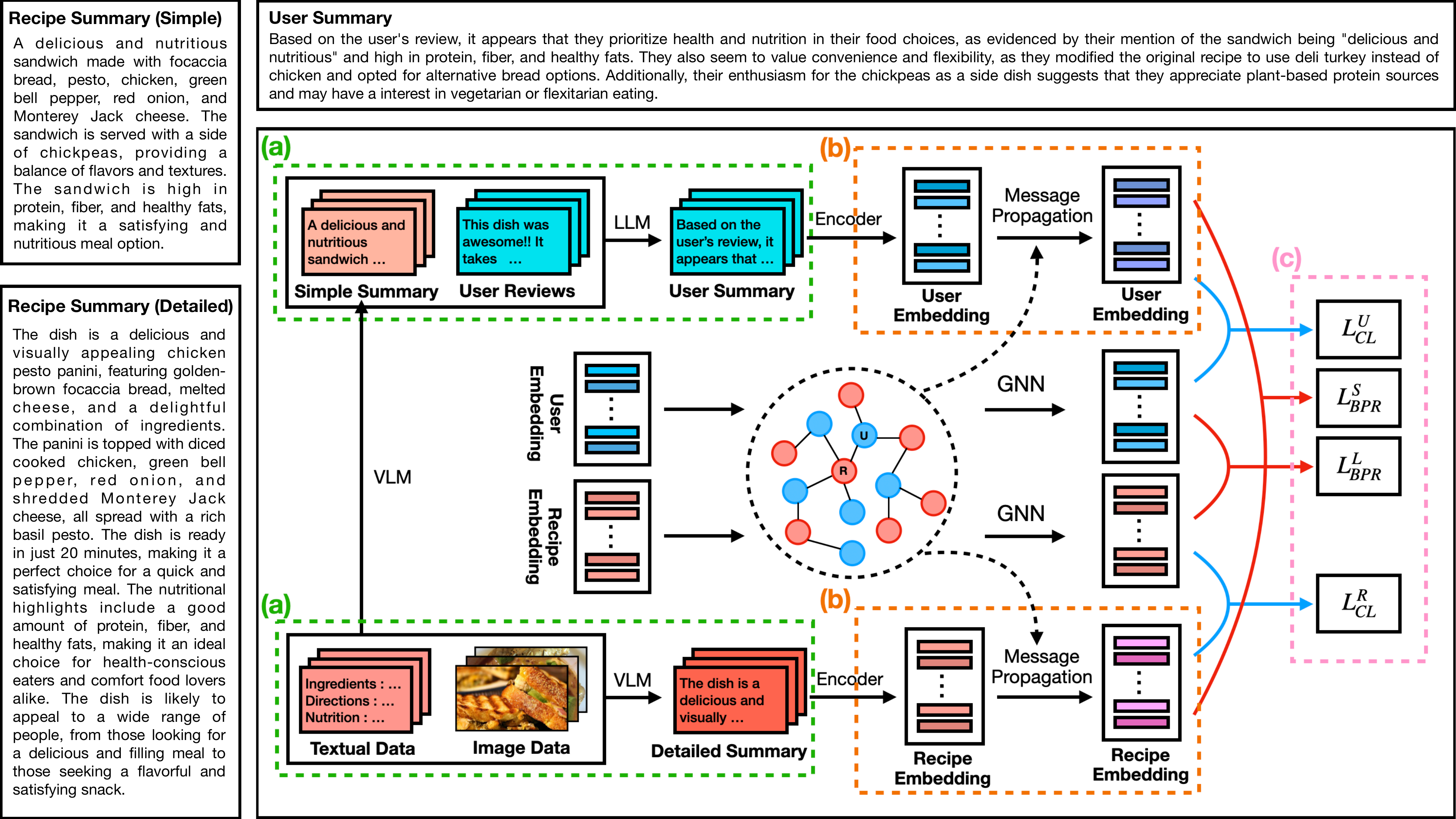}
        \caption{Overview of \method with (a) content-, (b) relation-, and (c) learning-based enhancement of multimodal features.
        \label{fig:model}}
\end{figure*}

In this section, we present \method, our recipe recommender system that applies three stages (\textbf{T1}–\textbf{T3}) of multimodal feature enhancement. See Figure~\ref{fig:model} for an overview.

\smallsection{(T1) Content-based enhancement.} 
To reduce noise and extract key information, we use pretrained foundation models with strong contextual understanding to generate summaries for recipes and users, as shown in part (a) of Figure~\ref{fig:model}.
Specifically, we apply a vision–language model (VLM)\footnote{\url{https://huggingface.co/llava-hf/llava-1.5-7b-hf}}
 to transform recipe features (titles, ingredients, directions, nutrition, and images) into (1) a \textit{simple summary} of all features except directions, and (2) a \textit{detailed summary} of all features, as follows:
$S_r^{\text{simple}}, S_r^{\text{detailed}} = \text{VLM}(\mathbf{V}_{r}, \mathbf{T}_{r})$.
For users, we use an LLM\footnote{\url{https://huggingface.co/meta-llama/Llama-3.1-8B-Instruct}}
 to summarize user preferences from textual reviews, combining each review with the corresponding simple recipe summary to provide context, i.e.,
 $S_u = \text{LLM}\left( \left\{ (S_r^{\text{simple}}|| \mathbf{T}_{u}^{r}) : A_{ur}=1 \right\} \right)$, where $||$ is concatenation.
Note that the simple summaries of the recipes are used instead of their detailed summaries to keep the token count small.
 See Figure~\ref{fig:model} for examples of summaries.
 


\smallsection{(T2) Relation-based enhancement.}
Then, we further enrich these summaries by incorporating collaborative signals from the user-recipe interaction graph. This stage, illustrated in part (b) of Figure~\ref{fig:model}, begins by embedding the recipe and user summaries using a pretrained language model (PLM)\textsuperscript{3} as the encoder.
\begin{equation*}\label{eq:init_embed}
\mathbf{e}_u^{(0)} = \mathrm{PLM}(S_u), \quad \mathbf{e}_r^{(0)} = \mathrm{PLM}\!\left(S_r^{\mathrm{detailed}}\right).
\end{equation*}
While semantically rich, these embeddings do not yet capture collaborative signals, so we apply LightGCN message propagation.
At each layer $k$, user and recipe embeddings are updated as:
\begin{equation*} 
\mathbf{e}_u^{(k+1)} 
= \sum_{r \in \mathcal{N}_u}
    \frac{1}{\sqrt{|\mathcal{N}_u|}\sqrt{|\mathcal{N}_r|}}
    \,\mathbf{e}_r^{(k)}, \ 
\mathbf{e}_r^{(k+1)} 
= \sum_{u \in \mathcal{N}_r}
    \frac{1}{\sqrt{|\mathcal{N}_r|}\sqrt{|\mathcal{N}_u|}}
    \,\mathbf{e}_u^{(k)},
\end{equation*}
where $|\mathcal{N}_u|$ and $|\mathcal{N}_r|$ denote the numbers of neighbors for user $u$ and recipe $r$, respectively.
After propagation, we obtain final embeddings by averaging the embeddings across all layers as: 
\begin{equation}\label{eq:equation2}
\mathbf{e}_u^{S} 
= \frac{1}{K+1} \sum\nolimits_{k=0}^{K} \mathbf{e}_u^{(k)},
\quad
\mathbf{e}_r^{S} 
= \frac{1}{K+1} \sum\nolimits_{k=0}^{K} \mathbf{e}_r^{(k)}.
\end{equation}
Note that this step introduces no additional parameters or training.

\smallsection{(T3) Learning-based enhancement.}
After that, as illustrated in part (c) of Figure~\ref{fig:model}, we further refine these content-based embeddings by aligning them with learning-based embeddings, improving their reliability especially when the corresponding multimodal content is noisy or limited.
Specifically, we obtain the learning-based embeddings, denoted by $\mathbf{e}_u^{L}$ and $\mathbf{e}_r^{L}$, 
using LightGCN with learnable initial embeddings,
and perform cross-view contrastive learning between the two embedding types using the InfoNCE loss~\cite{oord2018InfoNCE}, i.e.,
\begin{align*}
    \mathcal{L}_{\text{CL}}^{\text{U}} &= - \sum_{u \in \mathcal{U}} \log \frac{\exp \left( \text{cos} \left( \mathbf{e}_u^{S}, \mathbf{e}_u^{L} \right) / \tau \right)}{\sum_{u' \in \mathcal{U}} \exp \left( \text{cos} \left( \mathbf{e}_u^{S}, {\mathbf{e}}_{u'}^{L} \right) / \tau \right)}, \end{align*}
where $\tau$ is a temperature hyperparameter and $\text{cos}(\cdot,\cdot)$ is cosine similarity. The recipe-side contrastive loss $\mathcal{L}_{\text{CL}}^{\text{R}}$ is defined likewise, and the final contrastive loss is the sum of the two components:
\begin{equation}\label{eq:equation_cl_total}
    \mathcal{L}_{\text{CL}} 
    = \mathcal{L}_{\text{CL}}^{\text{U}} + \mathcal{L}_{\text{CL}}^{\text{R}}.
\end{equation}

\smallsection{Training and scoring.}
For the training objective, we use the BPR loss~\cite{rendle2012bpr} for each embedding type, in addition to the above cross-view contrastive loss and a regularization loss as follows:
\begin{equation*}
    \mathcal{L}_{\text{total}} = \mathcal{L}_{\text{BPR}}^{\text{S}} + \mathcal{L}_{\text{BPR}}^{\text{L}} + \lambda_{CL} \mathcal{L}_{\text{CL}} + \lambda_{REG} (\sum_{r\in \mathcal{R}}||\mathbf{\bar{e}}^{(0)}_r||_2^2+\sum_{u\in \mathcal{U}}||\mathbf{\bar{e}}^{(0)}_u||_2^2),
\end{equation*}
where $\mathbf{\bar{e}}^{(0)}_r$ and $\mathbf{\bar{e}}^{(0)}_u$ are learnable embeddings of recipe $r$ and user $u$ (before LightGCN); and $\lambda_{CL}$ and $\lambda_{REG}$ are hyperparameters.

After training,
for each user–recipe pair $(u, r)$, the final score for recommendation is computed using inner products from both embedding types as follows: $(\mathbf{e}_r^{S})^\top\mathbf{e}_u^{S}+(\mathbf{e}_r^{L})^\top\mathbf{e}_u^{L}$.

\smallsection{Comparison with existing multimodal recommenders.} Ours leverages user reviews to generate user embeddings, while existing ones either overlook reviews entirely~\cite{dang2025MLLMRec, su2024SOIL, guo2025MMHCL} or treat them as item-level features, aggregating all reviews received by an item~\cite{liu2023MGCL, liu2023MEGCF}.
Moreover, we apply message propagation to both the content-based embeddings and the learnable embeddings and then contrast them, achieving more effective alignment between the two. In contrast, most existing methods apply message propagation only to the learnable embeddings before alignment~\cite{zhou2023BM3, tao2022slmrec}.
We empirically verify their substantial contributions to performance in RQ2 of Section~\ref{sec:experiments:results}  (by comparing \method-T2 and \method-R with \method).

\section{Experiments}
\label{sec:experiments}


In this section, we review our experiments for three questions:
\begin{itemize}[leftmargin=*]
    \item \textbf{RQ1. Performance comparison.} How effective is \method, compared to state-of-the-art recipe recommender systems?
    \item \textbf{RQ2. Ablation study.} Do all the key components of \method contribute significantly to its performance?
    \item \textbf{RQ3. Speed and memory efficiency.} How fast and memory-efficient is \method during training and inference?
    \item \textbf{RQ4. Hyperparameter effects.} How do hyperparameters affect the performance of \method?
    
\end{itemize}

\subsection{Experimental Settings}

\smallsection{Datasets and splits.}
We use the two benchmark datasets described above, following the preprocessing in~\cite{zhang2024clussl, zhang2024healthrec}. 

\smallsection{Baseline methods.}
For baselines, we use eight baseline methods, including 
 two general recommender systems (BPRMF~\cite{rendle2012bpr} and LightGCN~\cite{he2020lightgcn}), three multimodal recommender systems (BM3 \cite{zhou2023BM3}, DRAGON~\cite{zhou2023DRAGON}, and FREEDOM~\cite{zhou2023FREEDOM}) and three recipe recommender systems (FGCN~\cite{gao2022fgcn}, CLUSSL~\cite{zhang2024clussl}, and HealthRec~\cite{zhang2024healthrec}).

\smallsection{Evaluation metrics.}
We employ Recall@$k$ and NDCG@$k$ with $k\in\{10,20\}$ as in \cite{zhou2023BM3, zhou2023FREEDOM, zhou2023DRAGON, zhang2024clussl, zhang2024healthrec}.
We conduct five runs with different random seeds and report the average performance.


\smallsection{Hyperparameters.}
For all methods, we use the Xavier initialization~\cite{xavier2010understanding} and the Adam optimizer~\cite{kinga2015adam}. We fix the feature embedding dimension to 64, train batch size to 2048, and evaluation batch size to 4096. We fix the learning rate to 1e-3 and follow the configuration of the regularization weight $\lambda_{REG}$ used in the MMRec code.\footnotemark
For \method, we tune the LightGCN layers in $\{1, 2, 3\}$, the temperature $\tau$ in $\{0.3, 0.5, 0.7\}$, and the weight $\lambda_{CL}$ in $\{0.1, 0.2, 0.3, 0.5\}$.

\smallsection{Machines.}
All experiments were run on a server with four NVIDIA RTX A6000 GPUs and 512 GB RAM.



\subsection{Experimental Results}\label{sec:experiments:results}

\smallsection{(RQ1) Performance comparison.}
As shown in Table~\ref{tab:results}, \method outperforms all eight baselines across all the metrics and datasets, achieving a 7-15$\%$ performance gain in Recall@10 over the best baseline. 
Notably, the recipe recommender systems~\cite{gao2022fgcn, zhang2024clussl, zhang2024healthrec} struggle on the Allrecipes dataset. We attribute this to their reliance on the recipe-ingredient graph for learning recipe embeddings.
The ingredient-to-recipe ratio is 2.6$\times$ higher in Allrecipes than Food-.com, as shown in Table~\ref{tab:dataset}.
We hypothesize that recipes in Allrecipes contain weakly informative or overly generic ingredients, which introduce noise and make it difficult for graph-based methods to learn effective recipe embeddings.
In contrast, \method leverages the foundation model’s semantic understanding of ingredients, allowing it to emphasize meaningful ingredients in its generated summaries, which makes \method robust to noisy or generic ingredients.


\renewcommand{\arraystretch}{1.0}
\begin{table}[t!]
\vspace{-1mm}
\caption{\label{tab:results} 
Recipe recommendation performance. 
The best performance is highlighted in \textbf{bold}, and the second-best one is \underline{underlined}. 
 R@k: Recall@k. N@k: NDCG@k. Note that, 
  in all the cases, \method outperforms all the baseline methods.
}
\setlength\tabcolsep{1.5pt} 
\def\arraystretch{0.9}
\scalebox{0.83}{
\begin{tabular}{l|cccc|cccc}
    \toprule
    \textbf{Datasets} & \multicolumn{4}{c|}{\textbf{Allrecipes}} & \multicolumn{4}{c}{\textbf{Food.com}} \\
    \midrule
    \textbf{Metrics} & \textbf{R@10} & \textbf{N@10} & \textbf{R@20} & \textbf{N@20} & \textbf{R@10} & \textbf{N@10} & \textbf{R@20} & \textbf{N@20} \\
    \midrule
    \textbf{BPRMF}~\cite{rendle2012bpr} & 0.0201 & 0.0126 & 0.0433 & 0.0195 & 0.0236 & 0.0222 & 0.0414 & 0.0276 \\
    \textbf{LightGCN}~\cite{he2020lightgcn} & 0.0203 & 0.0138 & 0.0444 & 0.0206 & 0.0263 & 0.0241 & 0.0457 & 0.0300 \\
    \midrule
    \textbf{BM3}~\cite{zhou2023BM3} & 0.0220 & 0.0146 & \underline{0.0449} & \underline{0.0212} & \underline{0.0271} & \underline{0.0253} & 0.0465 & 0.0311 \\
    \textbf{DRAGON}~\cite{zhou2023DRAGON} & \underline{0.0234} & \underline{0.0152} & 0.0422 & 0.0206 & 0.0269 & 0.0248 & 0.0455 & 0.0302 \\
    \textbf{FREEDOM}~\cite{zhou2023FREEDOM} & 0.0224 & 0.0148 & 0.0441 & 0.0211 & 0.0264 & 0.0248 & 0.0445 & 0.0302 \\
    \midrule
    \textbf{FGCN}~\cite{gao2022fgcn} & 0.0202 & 0.0133 & 0.0448 & 0.0204 & 0.0250 & 0.0227 & 0.0452 & 0.0291 \\
    \textbf{CLUSSL}~\cite{zhang2024clussl} & 0.0225 & 0.0148 & 0.0432 & 0.0207 & 0.0263 & 0.0250 & 0.0461 & 0.0311 \\
    \textbf{HealthRec}~\cite{zhang2024healthrec} & 0.0213 & 0.0141 & 0.0432 & 0.0205 & 0.0266 & \underline{0.0253} & \underline{0.0469} & \underline{0.0314} \\
    \midrule
    \textbf{\method}   & \textbf{0.0270} &   \textbf{0.0166} &  \textbf{0.0484} &  \textbf{0.0228} & \ \textbf{0.0290} & \ \textbf{0.0272} &  \textbf{0.0490} &  \textbf{0.0330}  \\
    \rowcolor{verylightgray} \textbf{Improvement} & 15.38\% & 9.21\% & 7.80\% & 7.55\% & 7.01\% & 7.51\% & 4.48\% & 5.10\% \\
    \bottomrule
\end{tabular}}
\end{table}

{\renewcommand{\arraystretch}{1.0}
\begin{table}[t!]
\vspace{-2mm}
\caption{\label{tab:ablation} Effectiveness of the key components of \method. The best performance is highlighted in \textbf{bold}, and the second-best one is \underline{underlined}. R@k: Recall@k. N@k: NDCG@k.
}
\setlength\tabcolsep{4.pt} 
\scalebox{0.77}{
\begin{tabular}{l|cccc|cccc}
    \toprule
    \textbf{Datasets} & \multicolumn{4}{c|}{\textbf{Allrecipes}} & \multicolumn{4}{c}{\textbf{Food.com}} \\
    \midrule
    \textbf{Metrics} & \textbf{R@10} & \textbf{N@10} & \textbf{R@20} & \textbf{N@20} & \textbf{R@10} & \textbf{N@10} & \textbf{R@20} & \textbf{N@20} \\
    \midrule
    \textbf{\method-T1.} & \underline{0.0255} & \underline{0.0159} &  0.0479 & \underline{0.0225} & 0.0277 & 0.0261 & 0.0478 & 0.0321 \\
    \textbf{\method-T2.} & 0.0180 & 0.0121 & 0.0299 & 0.0155 & 0.0156 & 0.0157 & 0.0242 & 0.0178 \\ 
    \textbf{\method-T3.} & 0.0249 & 0.0155 & 0.0476 & \underline{0.0225} & 0.0281 & 0.0264 & \underline{0.0483} & 0.0324 \\
    \textbf{\method-R.} & 0.0242 & 0.0154 & \underline{0.0480} & 0.0223 & \underline{0.0284} & \underline{0.0269} & 0.0478 & \underline{0.0325}\\ 
    \textbf{\method-V.} & 0.0250 & 0.0158 & 0.0477 & \underline{0.0225} & 0.0259  & 0.0257 & 0.0469 & 0.0319 \\ 
    \midrule
    \textbf{\method} & \textbf{0.0270} & \textbf{0.0166} & \textbf{0.0484} & \textbf{0.0228} & \textbf{0.0290} & \textbf{0.0272} & \textbf{0.0490} & \textbf{0.0330} \\
    \bottomrule
\end{tabular}}
\end{table}
}

\footnotetext{\url{https://github.com/enoche/MMRec}}

\smallsection{(RQ2) Ablation study.}
To validate the contribution of each component in \method, we consider five variants of \method: (1) \textbf{\method-T1}: without content-based enhancement, 
(2) \textbf{\method-T2}: without relation-based enhancement, (3) \textbf{\method-T3}: without learning-based enhancement, (4) \textbf{\method-R}: without leveraging user reviews in user summary generation, and (5) \textbf{\method-V}: without leveraging recipe images in both simple and detailed recipe summary generation.
As shown in Table~\ref{tab:ablation}, \method outperforms its variants in all cases, which demonstrates the effectiveness of each key component. 
Notably, removing the relation-based enhancement (i.e., \method-T2) yields the largest performance drop, indicating that propagating content-based summaries over the user-recipe interaction graph, thereby fusing content-based and collaborative signals, is critical. Furthermore, without using user reviews or recipe images for summary generation, \method exhibits a performance drop, demonstrating the importance of using multimodal information.


\smallsection{(RQ3) Speed and memory efficiency.}
We compare the efficiency of \method with the three recipe recommender systems in four aspects: (1) the average training time per epoch, (2) the average time taken for the performance of model to converge, (3) the average time taken for inference over all users, and (4) the maximum amount of VRAM allocated during training.
For this evaluation, we use the Allrecipes dataset.
As shown in Table~\ref{tab:efficiency}, \method achieves better efficiency than the competing methods across all aspects. Notably, \method further requires an additional preprocessing step for user and recipe summarization, which takes approximately 9 hours and 5 hours, respectively, corresponding to about 0.5 seconds per user and 0.4 seconds per recipe.
However, it is performed only once as an offline preprocessing step and does not add overhead during model training or inference.

\begin{figure}[t]
    \vspace{-3mm}
    \centering 
\includegraphics[width=0.99\linewidth]{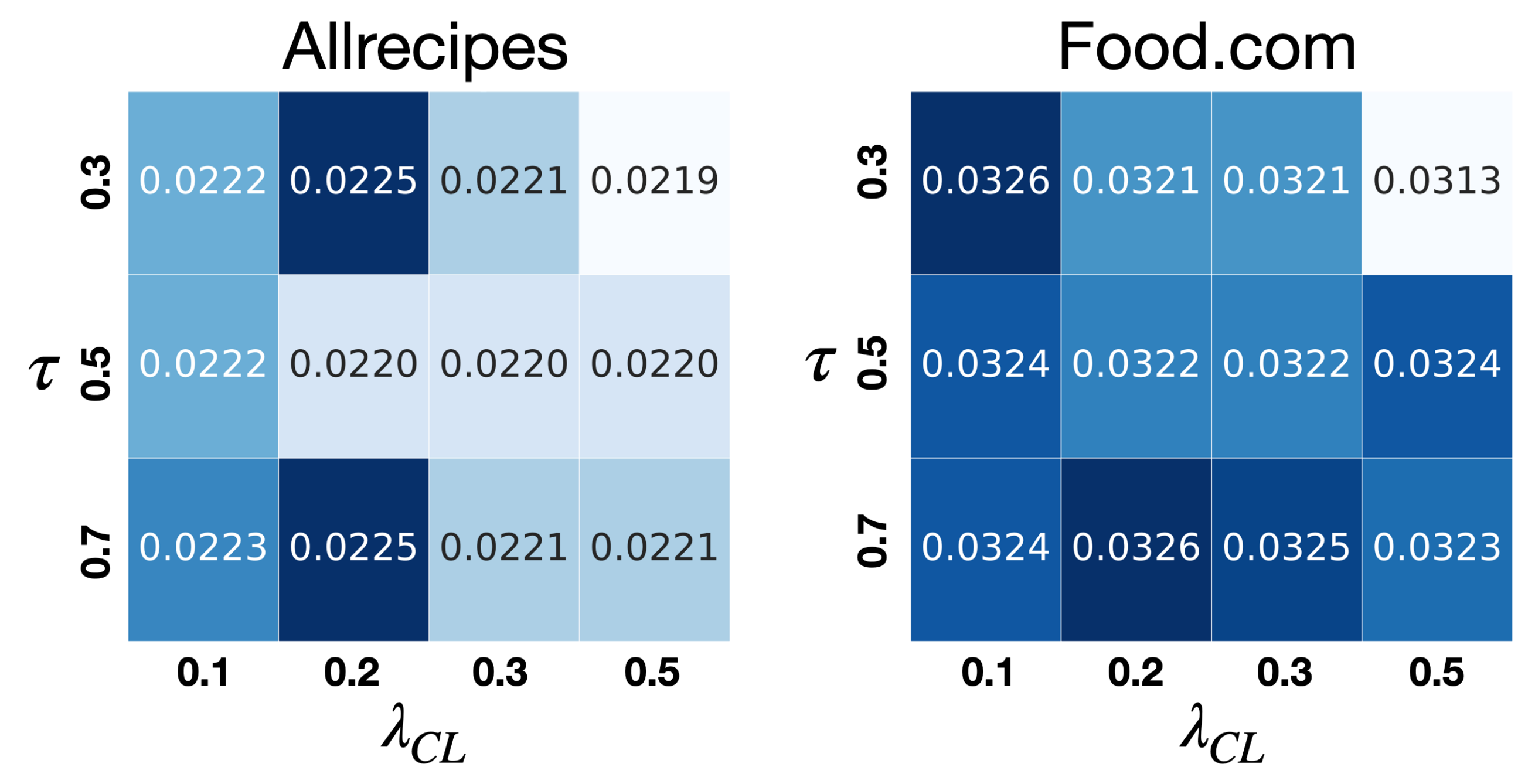}
        \caption{Effects of $\tau$ and $\lambda_{CL}$ on the NDCG@20 of \method. \label{fig:hyperparameter}
        } 
\end{figure}

\smallsection{(RQ4) Hyperparameter effects.}
We examine the effect of two key hyperparameters in \method: the contrastive-learning temperature and loss weight. 
As shown in Figure~\ref{fig:hyperparameter}, \method is robust to the choice of temperature and loss weight parameters, with minimal performance variation. Specifically, the maximum gap between the best and worst settings is 4.15\% in terms of NDCG@20.

{\renewcommand{\arraystretch}{1.0}
\begin{table}
\vspace{-1mm}
\caption{\label{tab:efficiency} 
Speed and memory efficiency of \method and recipe recommendation methods. The best performance is highlighted in \textbf{bold}, and the second-best one is \underline{underlined}.}

\setlength\tabcolsep{4.pt} 
\scalebox{0.8}{
\begin{tabular}{l|cccc}
    \toprule
    \textbf{Dataset} & \multicolumn{4}{c}{\textbf{Allrecipes}} \\
    \midrule
    \textbf{Metrics} & \textbf{Train} & \textbf{Converge} & \textbf{Inference} & \textbf{VRAM} \\
    \midrule
    \textbf{FGCN}~\cite{gao2022fgcn} & \underline{75.13s} & 47min & 16.44s  & \underline{13077MB}\\
    \textbf{CLUSSL}~\cite{zhang2024clussl} & 86.19s & \underline{36min} & \underline{13.62s} & 13107MB  \\ 
    \textbf{HealthRec}~\cite{zhang2024healthrec} & 107.31s & 55min & 15.45s & 15404MB \\
    \midrule
    \textbf{\method} & \textbf{25.67s} & \textbf{15min} & \textbf{5.72s} & \textbf{4732MB}\\
    \bottomrule
\end{tabular}}
\end{table}
}

\section{Conclusions and Future Directions}
\label{sec:conclusion}

In this work, we introduce \method, a multimodal recipe recommender system motivated by our preliminary analysis, which demonstrates the critical role and strong potential of leveraging multimodal features.
\method improves these features through content-, relation-, and learning-based enhancements; employing foundation models, message propagation, and contrastive learning, respectively.
Experiments on real-world datasets show that \method consistently outperforms eight baseline methods, achieving a 7–15\% improvement in Recall@10. Our code and datasets are available at \url{https://github.com/JHshin6688/TESMR}.

As future work, our work can be extended toward end-to-end optimization that jointly learns foundation-model-based feature enhancement and graph-based recommendation models. Another possible direction is to model the temporal evolution of multimodal user preferences, moving beyond static user summaries.

\section*{Acknowledgements}
This work was partly supported by the National Research Foundation of Korea (NRF) grant funded by the Korea government (MSIT) (No. RS-2024-00406985, 40\%).
This work was partly supported by Institute of Information \& Communications Technology Planning \& Evaluation (IITP) grant funded by the Korea government (MSIT) (No. RS-2024-00438638, EntireDB2AI: Foundations and Software for Comprehensive Deep Representation Learning and Prediction on Entire Relational Databases, 50\%)
(No. RS-2019-II190075, Artificial Intelligence Graduate School Program (KAIST), 10\%).
        
\bibliographystyle{ACM-Reference-Format}
\bibliography{ref}


\begin{thebibliography}{22}


\ifx \showCODEN    \undefined \def \showCODEN     #1{\unskip}     \fi
\ifx \showDOI      \undefined \def \showDOI       #1{#1}\fi
\ifx \showISBNx    \undefined \def \showISBNx     #1{\unskip}     \fi
\ifx \showISBNxiii \undefined \def \showISBNxiii  #1{\unskip}     \fi
\ifx \showISSN     \undefined \def \showISSN      #1{\unskip}     \fi
\ifx \showLCCN     \undefined \def \showLCCN      #1{\unskip}     \fi
\ifx \shownote     \undefined \def \shownote      #1{#1}          \fi
\ifx \showarticletitle \undefined \def \showarticletitle #1{#1}   \fi
\ifx \showURL      \undefined \def \showURL       {\relax}        \fi
\providecommand\bibfield[2]{#2}
\providecommand\bibinfo[2]{#2}
\providecommand\natexlab[1]{#1}
\providecommand\showeprint[2][]{arXiv:#2}

\bibitem[Dang et~al\mbox{.}(2025)]%
        {dang2025MLLMRec}
\bibfield{author}{\bibinfo{person}{Yuzhuo Dang}, \bibinfo{person}{Xin Zhang}, \bibinfo{person}{Zhiqiang Pan}, \bibinfo{person}{Yuxiao Duan}, \bibinfo{person}{Wanyu Chen}, \bibinfo{person}{Fei Cai}, {and} \bibinfo{person}{Honghui Chen}.} \bibinfo{year}{2025}\natexlab{}.
\newblock \showarticletitle{MLLMRec: Exploring the Potential of Multimodal Large Language Models in Recommender Systems}.
\newblock \bibinfo{journal}{\emph{arXiv:2508.15304}} (\bibinfo{year}{2025}).
\newblock


\bibitem[Fu et~al\mbox{.}(2025)]%
        {fu2025ChefMind}
\bibfield{author}{\bibinfo{person}{Yu Fu}, \bibinfo{person}{Linyue Cai}, \bibinfo{person}{Ruoyu Wu}, {and} \bibinfo{person}{Yong Zhao}.} \bibinfo{year}{2025}\natexlab{}.
\newblock \showarticletitle{From" What to Eat?" to Perfect Recipe: ChefMind's Chain-of-Exploration for Ambiguous User Intent in Recipe Recommendation}.
\newblock \bibinfo{journal}{\emph{arXiv:2509.18226}} (\bibinfo{year}{2025}).
\newblock


\bibitem[Gao et~al\mbox{.}(2022)]%
        {gao2022fgcn}
\bibfield{author}{\bibinfo{person}{Xiaoyan Gao}, \bibinfo{person}{Fuli Feng}, \bibinfo{person}{Heyan Huang}, \bibinfo{person}{Xian-Ling Mao}, \bibinfo{person}{Tian Lan}, {and} \bibinfo{person}{Zewen Chi}.} \bibinfo{year}{2022}\natexlab{}.
\newblock \showarticletitle{Food recommendation with graph convolutional network}.
\newblock \bibinfo{journal}{\emph{Information Sciences}}  \bibinfo{volume}{584} (\bibinfo{year}{2022}), \bibinfo{pages}{170--183}.
\newblock


\bibitem[Glorot and Bengio(2010)]%
        {xavier2010understanding}
\bibfield{author}{\bibinfo{person}{Xavier Glorot} {and} \bibinfo{person}{Yoshua Bengio}.} \bibinfo{year}{2010}\natexlab{}.
\newblock \showarticletitle{Understanding the difficulty of training deep feedforward neural networks}. In \bibinfo{booktitle}{\emph{AISTATS}}.
\newblock


\bibitem[Guo et~al\mbox{.}(2025)]%
        {guo2025MMHCL}
\bibfield{author}{\bibinfo{person}{Xu Guo}, \bibinfo{person}{Tong Zhang}, \bibinfo{person}{Fuyun Wang}, \bibinfo{person}{Xudong Wang}, \bibinfo{person}{Xiaoya Zhang}, \bibinfo{person}{Xin Liu}, {and} \bibinfo{person}{Zhen Cui}.} \bibinfo{year}{2025}\natexlab{}.
\newblock \showarticletitle{MMHCL: Multi-Modal Hypergraph Contrastive Learning for Recommendation}.
\newblock \bibinfo{journal}{\emph{ACM TOMM}} \bibinfo{volume}{21}, \bibinfo{number}{10} (\bibinfo{year}{2025}), \bibinfo{pages}{1--23}.
\newblock


\bibitem[He et~al\mbox{.}(2020)]%
        {he2020lightgcn}
\bibfield{author}{\bibinfo{person}{Xiangnan He}, \bibinfo{person}{Kuan Deng}, \bibinfo{person}{Xiang Wang}, \bibinfo{person}{Yan Li}, \bibinfo{person}{Yongdong Zhang}, {and} \bibinfo{person}{Meng Wang}.} \bibinfo{year}{2020}\natexlab{}.
\newblock \showarticletitle{Lightgcn: Simplifying and powering graph convolution network for recommendation}. In \bibinfo{booktitle}{\emph{SIGIR}}.
\newblock


\bibitem[Kinga et~al\mbox{.}(2015)]%
        {kinga2015adam}
\bibfield{author}{\bibinfo{person}{Diederik Kinga}, \bibinfo{person}{Jimmy~Ba Adam}, {et~al\mbox{.}}} \bibinfo{year}{2015}\natexlab{}.
\newblock \showarticletitle{A method for stochastic optimization}. In \bibinfo{booktitle}{\emph{ICLR}}.
\newblock


\bibitem[Liu et~al\mbox{.}(2023a)]%
        {liu2023MGCL}
\bibfield{author}{\bibinfo{person}{Kang Liu}, \bibinfo{person}{Feng Xue}, \bibinfo{person}{Dan Guo}, \bibinfo{person}{Peijie Sun}, \bibinfo{person}{Shengsheng Qian}, {and} \bibinfo{person}{Richang Hong}.} \bibinfo{year}{2023}\natexlab{a}.
\newblock \showarticletitle{Multimodal graph contrastive learning for multimedia-based recommendation}.
\newblock \bibinfo{journal}{\emph{IEEE Transactions on Multimedia}}  \bibinfo{volume}{25} (\bibinfo{year}{2023}), \bibinfo{pages}{9343--9355}.
\newblock


\bibitem[Liu et~al\mbox{.}(2023b)]%
        {liu2023MEGCF}
\bibfield{author}{\bibinfo{person}{Kang Liu}, \bibinfo{person}{Feng Xue}, \bibinfo{person}{Dan Guo}, \bibinfo{person}{Le Wu}, \bibinfo{person}{Shujie Li}, {and} \bibinfo{person}{Richang Hong}.} \bibinfo{year}{2023}\natexlab{b}.
\newblock \showarticletitle{MEGCF: Multimodal entity graph collaborative filtering for personalized recommendation}.
\newblock \bibinfo{journal}{\emph{ACM TOIS}} \bibinfo{volume}{41}, \bibinfo{number}{2} (\bibinfo{year}{2023}), \bibinfo{pages}{1--27}.
\newblock


\bibitem[Lopez-Avila and Du(2025)]%
        {lopez2025survey}
\bibfield{author}{\bibinfo{person}{Alejo Lopez-Avila} {and} \bibinfo{person}{Jinhua Du}.} \bibinfo{year}{2025}\natexlab{}.
\newblock \showarticletitle{A Survey on Large Language Models in Multimodal Recommender Systems}.
\newblock \bibinfo{journal}{\emph{arXiv preprint arXiv:2505.09777}} (\bibinfo{year}{2025}).
\newblock


\bibitem[Meng et~al\mbox{.}(2025)]%
        {meng2025doge}
\bibfield{author}{\bibinfo{person}{Fanshen Meng}, \bibinfo{person}{Zhenhua Meng}, \bibinfo{person}{Ru Jin}, \bibinfo{person}{Rongheng Lin}, {and} \bibinfo{person}{Budan Wu}.} \bibinfo{year}{2025}\natexlab{}.
\newblock \showarticletitle{DOGE: LLMs-Enhanced Hyper-Knowledge Graph Recommender for Multimodal Recommendation}. In \bibinfo{booktitle}{\emph{AAAI}}.
\newblock


\bibitem[Oord et~al\mbox{.}(2018)]%
        {oord2018InfoNCE}
\bibfield{author}{\bibinfo{person}{Aaron van~den Oord}, \bibinfo{person}{Yazhe Li}, {and} \bibinfo{person}{Oriol Vinyals}.} \bibinfo{year}{2018}\natexlab{}.
\newblock \showarticletitle{Representation learning with contrastive predictive coding}.
\newblock \bibinfo{journal}{\emph{arXiv:1807.03748}} (\bibinfo{year}{2018}).
\newblock


\bibitem[Rendle et~al\mbox{.}(2012)]%
        {rendle2012bpr}
\bibfield{author}{\bibinfo{person}{Steffen Rendle}, \bibinfo{person}{Christoph Freudenthaler}, \bibinfo{person}{Zeno Gantner}, {and} \bibinfo{person}{Lars Schmidt-Thieme}.} \bibinfo{year}{2012}\natexlab{}.
\newblock \showarticletitle{BPR: Bayesian personalized ranking from implicit feedback}.
\newblock \bibinfo{journal}{\emph{arXiv:1205.2618}} (\bibinfo{year}{2012}).
\newblock


\bibitem[Song et~al\mbox{.}(2023)]%
        {song2023SCHGN}
\bibfield{author}{\bibinfo{person}{Yaguang Song}, \bibinfo{person}{Xiaoshan Yang}, {and} \bibinfo{person}{Changsheng Xu}.} \bibinfo{year}{2023}\natexlab{}.
\newblock \showarticletitle{Self-supervised calorie-aware heterogeneous graph networks for food recommendation}.
\newblock \bibinfo{journal}{\emph{ACM TOMM}} \bibinfo{volume}{19}, \bibinfo{number}{1s} (\bibinfo{year}{2023}), \bibinfo{pages}{1--23}.
\newblock


\bibitem[Su et~al\mbox{.}(2024)]%
        {su2024SOIL}
\bibfield{author}{\bibinfo{person}{Hongzu Su}, \bibinfo{person}{Jingjing Li}, \bibinfo{person}{Fengling Li}, \bibinfo{person}{Ke Lu}, {and} \bibinfo{person}{Lei Zhu}.} \bibinfo{year}{2024}\natexlab{}.
\newblock \showarticletitle{SOIL: Contrastive Second-Order Interest Learning for Multimodal Recommendation}. In \bibinfo{booktitle}{\emph{MM}}.
\newblock


\bibitem[Tao et~al\mbox{.}(2022)]%
        {tao2022slmrec}
\bibfield{author}{\bibinfo{person}{Zhulin Tao}, \bibinfo{person}{Xiaohao Liu}, \bibinfo{person}{Yewei Xia}, \bibinfo{person}{Xiang Wang}, \bibinfo{person}{Lifang Yang}, \bibinfo{person}{Xianglin Huang}, {and} \bibinfo{person}{Tat-Seng Chua}.} \bibinfo{year}{2022}\natexlab{}.
\newblock \showarticletitle{Self-supervised learning for multimedia recommendation}.
\newblock \bibinfo{journal}{\emph{IEEE Transactions on Multimedia}}  \bibinfo{volume}{25} (\bibinfo{year}{2022}), \bibinfo{pages}{5107--5116}.
\newblock


\bibitem[Zhang et~al\mbox{.}(2024a)]%
        {zhang2024clussl}
\bibfield{author}{\bibinfo{person}{Yixin Zhang}, \bibinfo{person}{Xin Zhou}, \bibinfo{person}{Qianwen Meng}, \bibinfo{person}{Fanglin Zhu}, \bibinfo{person}{Yonghui Xu}, \bibinfo{person}{Zhiqi Shen}, {and} \bibinfo{person}{Lizhen Cui}.} \bibinfo{year}{2024}\natexlab{a}.
\newblock \showarticletitle{Multi-modal food recommendation using clustering and self-supervised learning}. In \bibinfo{booktitle}{\emph{PRICAI}}.
\newblock


\bibitem[Zhang et~al\mbox{.}(2024b)]%
        {zhang2024healthrec}
\bibfield{author}{\bibinfo{person}{Yixin Zhang}, \bibinfo{person}{Xin Zhou}, \bibinfo{person}{Fanglin Zhu}, \bibinfo{person}{Ning Liu}, \bibinfo{person}{Wei Guo}, \bibinfo{person}{Yonghui Xu}, \bibinfo{person}{Zhiqi Shen}, {and} \bibinfo{person}{Lizhen Cui}.} \bibinfo{year}{2024}\natexlab{b}.
\newblock \showarticletitle{Multi-modal food recommendation with health-aware knowledge distillation}. In \bibinfo{booktitle}{\emph{CIKM}}.
\newblock


\bibitem[Zhang et~al\mbox{.}(2025)]%
        {zhang2025MOPI-HFRS}
\bibfield{author}{\bibinfo{person}{Zheyuan Zhang}, \bibinfo{person}{Zehong Wang}, \bibinfo{person}{Tianyi Ma}, \bibinfo{person}{Varun~Sameer Taneja}, \bibinfo{person}{Sofia Nelson}, \bibinfo{person}{Nhi Ha~Lan Le}, \bibinfo{person}{Keerthiram Murugesan}, \bibinfo{person}{Mingxuan Ju}, \bibinfo{person}{Nitesh~V Chawla}, \bibinfo{person}{Chuxu Zhang}, {et~al\mbox{.}}} \bibinfo{year}{2025}\natexlab{}.
\newblock \showarticletitle{Mopi-hfrs: A multi-objective personalized health-aware food recommendation system with llm-enhanced interpretation}. In \bibinfo{booktitle}{\emph{KDD}}.
\newblock


\bibitem[Zhou et~al\mbox{.}(2023b)]%
        {zhou2023DRAGON}
\bibfield{author}{\bibinfo{person}{Hongyu Zhou}, \bibinfo{person}{Xin Zhou}, \bibinfo{person}{Lingzi Zhang}, {and} \bibinfo{person}{Zhiqi Shen}.} \bibinfo{year}{2023}\natexlab{b}.
\newblock \showarticletitle{Enhancing dyadic relations with homogeneous graphs for multimodal recommendation}.
\newblock \bibinfo{journal}{\emph{arXiv:2301.12097}} (\bibinfo{year}{2023}).
\newblock


\bibitem[Zhou and Shen(2023)]%
        {zhou2023FREEDOM}
\bibfield{author}{\bibinfo{person}{Xin Zhou} {and} \bibinfo{person}{Zhiqi Shen}.} \bibinfo{year}{2023}\natexlab{}.
\newblock \showarticletitle{A tale of two graphs: Freezing and denoising graph structures for multimodal recommendation}. In \bibinfo{booktitle}{\emph{MM}}.
\newblock


\bibitem[Zhou et~al\mbox{.}(2023a)]%
        {zhou2023BM3}
\bibfield{author}{\bibinfo{person}{Xin Zhou}, \bibinfo{person}{Hongyu Zhou}, \bibinfo{person}{Yong Liu}, \bibinfo{person}{Zhiwei Zeng}, \bibinfo{person}{Chunyan Miao}, \bibinfo{person}{Pengwei Wang}, \bibinfo{person}{Yuan You}, {and} \bibinfo{person}{Feijun Jiang}.} \bibinfo{year}{2023}\natexlab{a}.
\newblock \showarticletitle{Bootstrap latent representations for multi-modal recommendation}. In \bibinfo{booktitle}{\emph{WWW}}.
\newblock


\end{thebibliography}


\end{document}